# ESTIMATION OF CLASSROOMS OCCUPANCY USING A MULTI-LAYER PERCEPTRON

**Eugénio Rodrigues[1]\*, Luísa Dias Pereira[1],
Adélio Rodrigues Gaspar[1], Álvaro Gomes[2], Manuel Carlos Gameiro da Silva[1]**

1: ADAI, LAETA, Department of Mechanical Engineering
Faculty of Sciences and Technology
University of Coimbra
Rua Luís Reis Santos, Pólo II, 3030-788 Coimbra, Portugal
\* e-mail: eugenio.rodrigues@gmail.com, web: http://www.adai.pt

2: INESC Coimbra, Department of Electrical and Computer Engineering
Faculty of Sciences and Technology
University of Coimbra
Rua Luís Reis Santos, Pólo II, 3030-290 Coimbra, Portugal
web: http://www.inescc.pt

**Abstract** *This paper presents a multi-layer perceptron model for the estimation of classrooms number of occupants from sensed indoor environmental data–relative humidity, air temperature, and carbon dioxide concentration. The modelling datasets were collected from two classrooms in the Secondary School of Pombal, Portugal. The number of occupants and occupation periods were obtained from class attendance reports. However, post-class occupancy was unknown and the developed model is used to reconstruct the classrooms occupancy by filling the unreported periods. Different model structure and environment variables combination were tested. The model with best accuracy had as input vector 10 variables of five averaged time intervals of relative humidity and carbon dioxide concentration. The model presented a mean square error of 1.99, coefficient of determination of 0.96 with a significance of p-value < 0.001, and a mean absolute error of 1 occupant. These results show promising estimation capabilities in uncertain indoor environment conditions.*

## 1. INTRODUCTION

Real-time occupancy data, by smart environmental control strategies, is definitely important to achieve energy-savings [1]. The importance of occupancy information for climate-control in buildings has been studied by different researchers [2]. Since measuring occupancy from sensors is not always feasible, as in commercial buildings, Liao and Barooah [3] developed an integrated approach (an agent-based model) "to simulate the behaviour of all the occupants of a building, and extract reduced-order graphical models from Monte-Carlo simulations of the agent-based model". Different approaches, such as "an information technology enabled sustainability test-bed" [4], RFID based system [5], electricity consumption [6] and other different sensing strategies [7] [8] have been applied for the same purpose: space occupancy detection. Other authors tackled this issue using gas sensors/concentrations, mostly $CO_2$, for occupancy estimations [9] [10]. Neural network models, have been less explored for this purpose [11] [12]. By enlarging the literature on space occupancy estimation, which as a fundamental role on indoor environmental control (IEC) decisions (either on energy simulation models or buildings in-use), and consequently on energy expenditures, this paper aims at contributing towards the challenge of modelling and estimation of occupancy, and therefore strengthening IEC decisions, ultimately improving indoor environmental quality conditions indoors and/or saving energy in buildings.

## 2. METHODOLOGY

### 2.1. Dataset

The SD800 Datalogger by Extech recorded indoor environment data–relative humidity (RH), air temperature ($T_{in}$), and carbon dioxide ($CO_2$) concentration—between 3 and 16 of April 2013 in two classrooms (classrooms 2.04 and 2.10) at the Secondary School in Pombal, Portugal. The measurements were taken in intervals of 60s. Due to normal class operation, the instrument was placed at a height of 2.70m above the floor in the middle of one of the rooms, thus not satisfying fully the ISO 7726 recommendations. The two classrooms have similar area and volume–around 50m$^2$ and 141m$^3$, respectively. The main difference was that one had its external wall facing northwest and the other southeast. Both classrooms were provided with centralized systems of thermal energy production and air renewal was ensured by air handling units with heating and cooling coil. The attendance reports of students classes were obtained. However, post-class occupancy was unknown. Therefore, only the periods of confirmed occupancy, weekends, and night time comprehended between midnight and 7:00 AM were considered in the modelling dataset. Also, when the data presented rapid variations readings of $CO_2$ concentration that did not seem plausible, these were also disregarded in the modelling dataset.

### 2.2. Multi-layer perceptron model

The model used to reconstruct the occupancy in the classrooms was a multi-layer perceptron (MLP) [13] trained by a resilient back-propagation algorithm with weight backtracking (RPROP+) [14]. The MLP model is a interconnect network of units in a feedforward manner. This means that information flows only in a forward manner. The basic unit of the model is the neuron or node. The neuron sums the weighted signals from the previous layer of neurons and transfers the new signal to the following layer according to an activation function—in this case a logistic sigmoid function. The first layer is the input layer where a vector of variables is used to predict an output. The MLP model used in this case combines two of three types of environment variables (relative humidity, air temperature, and carbon dioxide concentration) that correspond to the average of 5 intervals— [$t$-30, $t$-21], [$t$-20, $t$-3], [$t$-2, $t$+2], [$t$+3, $t$+20], and [$t$+21, $t$+30]. The time step is one minute. Thus, the model totalizes 10 input variables, five per each environment variable. The model output is the number of occupants (students plus one teacher). The model was implemented in R software and the *neuralnet* package was used.

Several model structures were tested to determine the one with the best accuracy–one and two hidden layers were used with different number of neurons. To achieve this, a fast 10-fold cross validation was used where the dataset was randomly split into 10 parts. Nine were used for training and one for validation. The models were trained with a threshold of 0.3 and the accuracy was measured by mean square error (MSE) indicator.

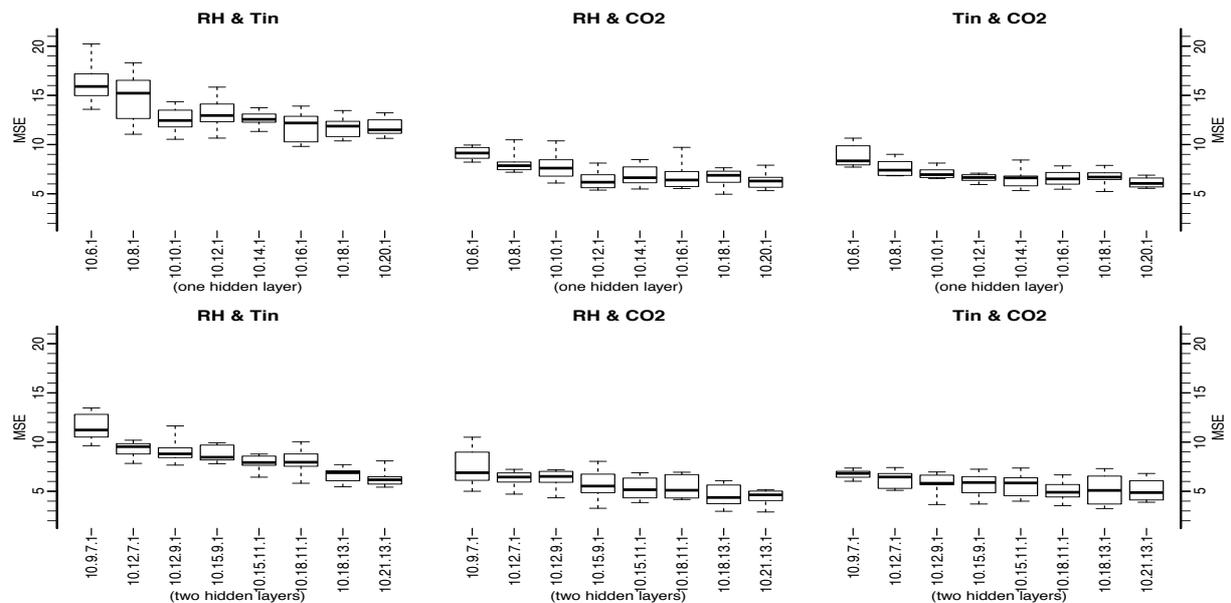

Figure 1. 10-fold cross validation of different MLP model structures. In the first row are depicted single hidden layer models with hidden neurons varying from 6 to 20. In the second row, two hidden layers' models are presented–neurons ranging between 9 and 21 in the first hidden layer, and 7 and 13 neurons in the second layer.

It is observable that the models with $CO_2$ input variables present the best accuracy and that the two hidden layer models are the most accurate ones. From all the model structures, the one that presents the best prediction capabilities was the RH & $CO_2$ model with 18 and 13 neurons in the hidden layers. Thus, this model was chosen to reconstruct the occupancy in the classrooms.

## 3. RESULTS AND DISCUSSION

After determining the structure of the MLP model, this was trained with a threshold of 0.03. 75% of the dataset was used for training and the remaining 25% for validation. When concluded the training, the model shown an accuracy in the validation set of MSE=1.99 occupants, coefficient of determination $R^2$=0.96 with a significance of p-value < 0.001, and a mean absolute error MAE=1 occupant. Figure 2 depicts the $CO_2$ and RH, as well as the reported and MLP estimated occupancy for classroom 2.04. It is possible to observe that the estimation values tend to follow the reported values and at the same time predict post-occupancy periods that are not in the attendance reports. However, there is also the tendency to underestimate the number of occupants immediately after the room is occupancy ends, thus predicting negative values. These changeable predictions may be justified with the fact that spaces are not fully occupied when the class starts and vacant when it ends, momentary absence of some students, manual opening of windows, manual control of HVAC system, and variation of the students' metabolism. However, and despite the results do not fully agree with the reported occupancy, these encourage future developments such as to include air temperature, fresh air flow, infiltration rates, wind speed and direction, exterior wall orientation, mean occupants age, and mean occupants activity levels, and so as to test other predictive methods.

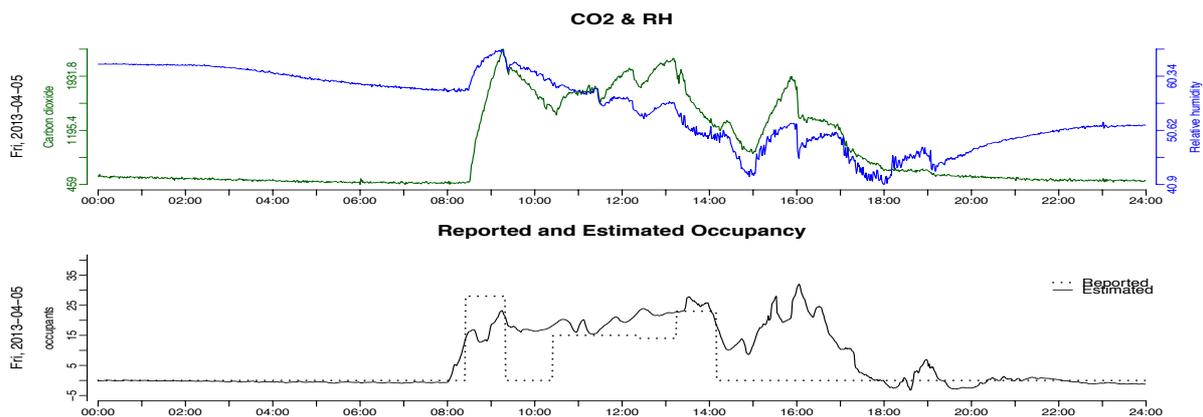

Figure 2. Secondary Pombal school classroom 2.04: CO2 and RH measurements (top graphic), and reported and estimated occupancy (bottom graphic).

## 4. CONCLUSION

Occupancy has a significant influence in the energy and indoor air quality performance of a space. The use of predictive tools to reconstruct the number and periods of occupancy of past sensed data is useful for a better understanding of the building behaviour and for the development of model predict control tools. Herein, several model structures were tested to reconstruct the occupancy in secondary schools' classrooms. The models with $CO_2$ input variables presented the best accuracy, and so, the chosen model – the one presenting the best prediction capabilities – was the RH & $CO_2$ model with 18 and 13 neurons in the hidden layers.

All in all, this work presented the preliminary work on the use of a MLP model with satisfactory accuracy in determining the number of occupants in two classrooms of a Secondary school. However, future work is necessary to improve the overall performance of the model and to compare it with other predictive methods.

**ACKNOWLEDGEMENTS**

The presented work is framed under the *Energy for Sustainability Initiative* of the University of Coimbra (UC). This work has been supported by the Portuguese Foundation for Science and Technology (FCT) and European Regional Development Fund (FEDER) through COMPETE—Programa Operacional Competitividade e Internacionalização (POCI) under the project Ren4EEnIEQ (PTDC/SEM-ENE/3238/2014 and POCI-01-0145-FEDER-016760 respectively); The first author acknowledges the support provided by the FCT under PostDoc grant SFRH/BPD/99668/2014.